\documentclass[runningheads]{llncs}
\usepackage[T1]{fontenc}
\usepackage{graphicx}
\usepackage{booktabs}
\usepackage[misc]{ifsym}

\usepackage{hyperref}
\usepackage{colortbl}
\usepackage{amsmath}
\usepackage{array}
\usepackage{amsfonts}
\newcommand{\vecX}{\mathbf{x}}
\newcommand{\vecP}{\mathbf{p}}
\newcommand{\vecH}{\mathbf{h}}

\begin{document}

\title{ALIEN: Aligned Entropy Head for Improving Uncertainty Estimation of LLMs}

\titlerunning{ALIEN: Aligned Entropy Head for Improving Uncertainty Estimation of LLMs}                    
  \author{Artem Zabolotnyi\inst{1} \and
  Roman Makarov\inst{1} \and
  Mile Mitrovic\inst{2} \and
  Polina Proskura\inst{3} \and
  Oleg Travkin\inst{1} \and
  Roman Alferov\inst{1} \and
  Alexey Zaytsev\inst{1}}


  \institute{Applied AI Institute, Moscow, Russia \and SB AI Lab, Moscow, Russia \and
  Intellectual data analysis and predictive modeling Institute, Moscow, Russia}

\maketitle              

\begin{abstract}
Uncertainty estimation remains a key challenge when adapting pre-trained language models to downstream classification tasks, with overconfidence often observed for difficult inputs. 
While predictive entropy provides a strong baseline for uncertainty estimation, it considers mainly aleatoric uncertainty and has limited capacity to capture effects, such as class overlap or ambiguous linguistic cues. 
We introduce Aligned Entropy - ALIEN, a lightweight method that refines entropy-based uncertainty by aligning it with prediction reliability. 
ALIEN trains a small uncertainty head initialized to produce the model’s original entropy and subsequently fine-tuned with two regularization mechanisms. 
Experiments across seven classification datasets and two NER benchmarks, evaluated on five language models (RoBERTa, ELECTRA, LLaMA-2, Qwen2.5, and Qwen3), show that ALIEN consistently outperforms strong baselines across all considered scenarios in detecting incorrect predictions, while achieving the lowest calibration error.
The proposed method introduces only a small inference overhead (in the order of milliseconds per batch on CPU) and increases the model’s parameter count by just ${\sim}0.002\%$ for decoder models and ${\sim}0.5\%$ for encoder models, without requiring storage of intermediate states. It improves uncertainty estimation while preserving the original model architecture, making the approach practical for large-scale deployment with modern language models.
Our results demonstrate that entropy can be effectively refined through lightweight supervised alignment, producing more reliable uncertainty estimates without modifying the backbone model.
The code is available at \footnote{\url{https://anonymous.4open.science/r/AdUE-1824/}}.

\end{abstract}

\section{Introduction}                                                             
Large language models are foundational in NLP, achieving strong transfer across tasks. 
For classification problems, a pretrained model is equipped with a head that transforms last-layer representations to class probabilities.
To improve the quality of representations, one considers fine-tuning methods that specifically train adapters~\cite{houlsby2019parameter}, including LoRA~\cite{hu2022lora}, especially for smaller models like BERT~\cite{liu2019roberta} or ELECTRA~\cite{clark2020electric}.
Parameter-efficient fine-tuning updates only a small subset of parameters, retaining most of the pre-trained backbone while enabling efficient adaptation with minimal compute and storage overhead.
Adapter-based models can solve a wide range of tasks, including risk-sensitive applications in medical~\cite{shool2025systematic,korchagin2025improving} and finance~\cite{fadeev2025latte} domains.

An accurate estimate of predictive
\emph{uncertainty} - that is, how confident the model is in its outputs, is therefore essential. 
Moreover, it directly translates into applications such as active learning~\cite{shelmanov2019active} and outlier detection~\cite{larson2019outlier}.
Common approaches~\cite{fadeeva2023lm} include entropy of the predictive distribution, Softmax Response~\cite{geifman2017selective},
Mahalanobis distance~\cite{NEURIPS2018_abdeb6f5}, and deep ensembles~\cite{lakshminarayanan2017simple}.
In practice, due to computational constraints, one opts for the first two methods, as they require little additional computational expense to obtain uncertainty.

However, both these options only partially cover the total uncertainty that consists of aleatoric and epistemic parts~\cite{hullermeier2021aleatoric}.
Aleatoric or data uncertainty relates to the intrinsic noise of the complexity of the data. 
Epistemic uncertainty refers to the model's lack of knowledge about the presented data.
Intrinsic predictive entropy of the answer provides a strong baseline, while covering only aleatoric uncertainty, while distance-based methods like Mahalanobis-distance-based~\cite{NEURIPS2018_abdeb6f5,van2020uncertainty} work better for capturing epistemic uncertainty. 
Vanilla entropy thus has a fundamental limitation, as it cannot account for all sources of uncertainty. 

We revisit predictive entropy as a lightweight yet effective uncertainty measure and propose ALIEN (ALigned ENtropy), a fine-tuning strategy to improve entropy-based uncertainty estimates. 
Our ALIEN operates on a frozen, task-tuned adapter backbone like a LoRA-adapted model and introduces a new uncertainty head initialized to reproduce the model's vanilla entropy scores. 
During training, ALIEN optimizes a multi-task loss with three components: (i) a binary cross-entropy term predicting whether the model’s prediction is incorrect (ii) a consistency regularization term to keep the new head's outputs close to the
original entropy scores, and (iii) an L2-SP anchoring term to keep the head's weights near their initialization~\cite{xuhong2018explicit}. 

Since we start from a natural entropy confidence score
(entropy), the fine-tuned uncertainty head should remain close to the original head. 
Moreover, while being relatively lightweight, classification heads often contain more parameters than there
are labeled training examples; thus, these regularizers are crucial to prevent overfitting.
Finally, observing actual model errors leads to a more complete understanding of the model's uncertainty. Thus, we allow the ALIEN head to better capture epistemic uncertainty, thereby improving overall model confidence estimates.

\begin{figure}[t]
  \centering
  \includegraphics[width=\textwidth]{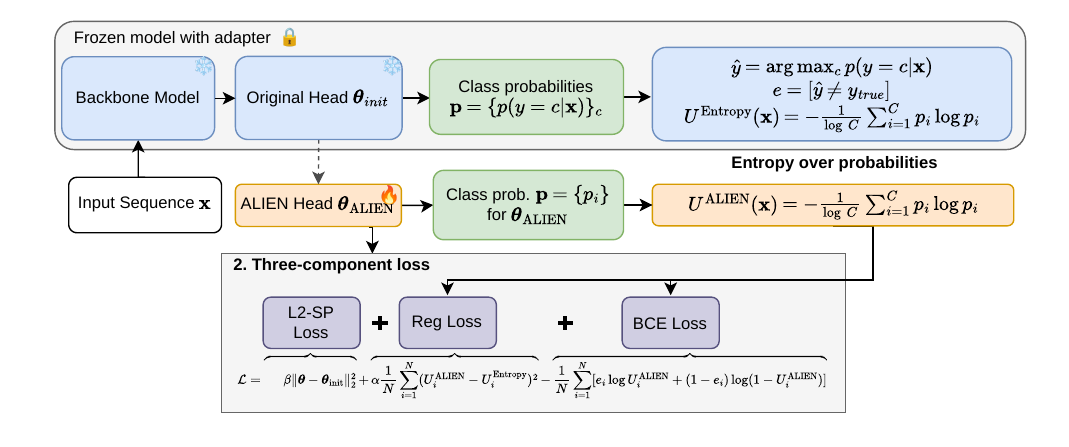}
  \caption{The ALIEN head training scheme. We initialize the new uncertainty head with the original weights $\theta_{init}$, use its entropy output as an initial uncertainty signal, and then
fine-tune the head with the three-term loss that includes binary cross-entropy, output consistency regularization, and L2-SP anchoring. The rest of the model (backbone and adapter) remains frozen.}
  \label{fig:ue_scheme}
\end{figure}

Our main contributions are thus as follows:
\begin{itemize}
\item We propose ALIEN, a lightweight uncertainty head that refines entropy-based uncertainty estimates for fine-tuned language models in a supervised way. Figure~\ref{fig:ue_scheme} illustrates the approach.
\item We introduce a training strategy combining classifier-head initialization, output-consistency regularization, and L2-SP anchoring to stabilize uncertainty refinement.
\item We show that learning from model errors enables ALIEN to capture uncertainty signals beyond vanilla predictive entropy, with stronger correlation with the epistemic component of uncertainty.
\item Extensive experiments across seven NLP classification benchmarks, two named entity recognition tasks, and five diverse language models show that ALIEN consistently improves uncertainty estimation compared to strong baselines, achieving better calibration while introducing negligible computational overhead.
\end{itemize}

\section{Related Work}
\label{sec:prior_work}
Predictive uncertainty in LLMs can be estimated using various methods. Among these, \emph{information-based} methods are widely utilized, which analyze token probability distributions by accessing logits or outputs from the internal layers of LLMs~\cite{takayama2019relevant,fomicheva2020unsupervised,van2022mutual,colombo2023rainproof}, or by relying on the generated text~\cite{tian2023just}. However, these methods are often outperformed by \emph{sample diversity} methods, which involve generating multiple outputs for LLM and either aggregating their confidence scores or assessing their diversity~\cite{kuhn2023semantic,malinin2020uncertainty,duan2023shifting}.

In contrast, \emph{density-based} methods approximate the distribution of training data using embeddings of training instances~\cite{NEURIPS2018_abdeb6f5,yoo-etal-2022-detection,ren2022out,vazhentsev2023hybrid}. Additionally, predictive uncertainty can be estimated using \emph{reflexive} methods, in which the model is asked directly to provide confidence levels for its responses~\cite{kadavath2022language,tian2023just}.

\section{Methods}

\subsection{Uncertainty estimation problem statement}
For a classification problem with labels $y$ from the set $\mathcal{C}$,
the model produces a vector of class probability estimates $\{ p(c \mid \vecX) \}_{c \in \mathcal{C}}$ with $\vecX$ being input, a natural language text in our context.
We aim to provide an uncertainty score $U(\vecX)$ that reflects the possibility of an error for an example $\vecX$.
The study focuses on computationally efficient methods, with our approach introducing only minimal overhead, comparable to applying the LLM’s classification layer to the output embedding. We use LoRA adapters for efficient, robust fine-tuning of an LLM for classification. 

After task fine-tuning, we freeze the backbone and adapter, so the only efficient option is to attach a head to the model's internal state $\vecH = h(\vecX) \in \mathbb{R}^d$. 
The vector $\vecH$ is typically extracted from the
penultimate layer of a neural network, while alternatives are possible~\cite{sky2024androids}.

\subsection{Probability-Based Uncertainty Estimates}

\textbf{Softmax Response (SR)} measures uncertainty via the complement of the maximum class probability~\cite{geifman2017selective}:
\begin{equation}
\label{eq:sr_formula}
U^{\mathrm{SR}}(\vecX) = 1 - \max_{c\in\mathcal{C}} p(c\mid \vecX),
\end{equation}
where $p(c\mid \vecX)$ are the softmax probabilities.

\textbf{Entropy} also quantifies uncertainty of the predictive distribution:
\begin{equation}
\label{eq:predictive-entropy}
U^{\mathrm{Entropy}}(\vecX) = -\frac{1}{\log C} \sum_{c=1}^{C} p(c \mid \vecX) \log p(c \mid \vecX).
\end{equation}
For binary classification, entropy provides a similar ranking to the SR ranking of samples.
For a multiclass setting, it exploits the full distribution and often provides superior results.

\subsection{ALIEN: Probability-Entropy Fine-Tuning}                   
While entropy often correlates with prediction reliability, it is not explicitly optimized to detect model errors. As a result, entropy may remain poorly aligned with actual model correctness. We therefore propose to learn a refined uncertainty signal that remains anchored to entropy but is trained to predict model mistakes. This allows the model to retain the strong baseline properties of entropy while incorporating additional signals reflecting epistemic uncertainty.

We introduce an ALIEN head which initially produces the exact same entropy scores~\eqref{eq:predictive-entropy} as the original model, as it is a duplicate of the trained final classifier layer.
After further alignment, the ALIEN head returns scores that, for each sample, more accurately predict the probability that the model's response is incorrect.

Let $\mathrm{ALIEN}(h(\vecX))$ denote the logits produced by the new uncertainty head for input $\vecX$. We obtain class probabilities via softmax:
\begin{equation}
  p(c \mid \vecX) = \mathrm{softmax}(\mathrm{ALIEN}(h(\vecX)))_c,
\end{equation}
for each class $c$. We initialize the head's parameters as $\theta_{\mathrm{ALIEN}} \leftarrow \theta_{\mathrm{init}}$ (the original classifier's weights) so that initially $p(c \mid \vecX)$ matches the
original classifier's outputs.

We define the ALIEN uncertainty score as the normalized entropy of this predictive distribution:
\begin{equation}
  U_{\mathrm{ALIEN}}(\vecX) =
  -\frac{1}{\log C}\sum_{c=1}^C p(c \mid \vecX) \log p(c \mid \vecX),
\end{equation}
which is bounded between 0 and 1.
By normalizing by $\log C$, we ensure that $U_{\mathrm{ALIEN}}(\vecX) = 1$ corresponds to maximum uncertainty with uniform distribution and $0$ to minimum uncertainty with a confident one-class output.

\subsection{ALIEN loss function}

The ALIEN head $U^{\mathrm{ALIEN}}(\cdot)$ with parameters $\boldsymbol{\theta}_{\mathrm{ALIEN}}$ is trained using an objective that combines three components, which incorporates the model error indicator $e_i = [\hat{y} \ne
y_{\mathrm{true}}]$, uncertainty estimates $U^{\mathrm{Entropy}}(\vecX)$, and parameters $\boldsymbol{\theta}_{\mathrm{init}}$ from the original classification head.

The first is a classic binary cross-entropy loss:
\[
  \mathcal{L}_{\mathrm{BCE}} = -\frac{1}{N}\sum_{i=1}^N \Bigl[e_i \log U_{\mathrm{ALIEN}} (\vecX_i) +
  (1 - e_i) \log(1-U_{\mathrm{ALIEN}}(\vecX_i)) \Bigr].
\]
The next term is a regularization loss:
\[
  \mathcal{L}_{\mathrm{reg}} = \frac{1}{N}\sum_{i=1}^N \left(U^{\mathrm{ALIEN}}(\vecX_i) - U^{\mathrm{Entropy}}(\vecX_i)\right)^2,
\]
which encourages the new uncertainty scores to stay close to the original entropy confidence, preventing drastic shifts.
Finally, we add L2-SP loss from the transfer learning domain~\cite{xuhong2018explicit}, which keeps the fine-tuned weights near their initialization, avoiding forgetting and overfitting:
\[
  \mathcal{L}_{\mathrm{L2SP}} = \|\boldsymbol{\theta}_{\mathrm{ALIEN}} - \boldsymbol{\theta}_{\mathrm{init}}\|_2^2.
\]

Our full training objective is a weighted sum of these components:
\[
  \mathcal{L} = \mathcal{L}_{\mathrm{BCE}} + \alpha \mathcal{L}_{\mathrm{reg}} + \beta \mathcal{L}_{\mathrm{L2SP}},
\label{eq:full_loss}
\]
where $\alpha$ and $\beta$ are hyperparameters.

\section{Experiments}

\subsection{Evaluation Protocol}

We evaluate uncertainty estimation by how well it identifies trustworthy predictions. We report ROC--AUC for separating correct from incorrect predictions~\cite{wang2024conuconformaluncertaintylarge}, AURC for the risk--coverage trade-off~\cite{geifman2017selective}, and ECE for calibration quality, where lower values indicate better agreement between confidence and accuracy~\cite{guo2017calibration}.

We follow an established protocol~\cite{fadeeva2024fact} and evaluate on seven text-classification datasets (SST-5, CoLA, 20 Newsgroups, ToxiGen, Yelp, GoEmotions, IMDB) and two token-level sequence-labeling (NER) benchmarks (CoNLL-2003, WNUT-2017).
See details in Table~\ref{table:datasets_sizes} and in the following paragraphs.

\begin{table*}[h!]
\caption{Summary of datasets used in our evaluation. The table reports the task type, number of classes, and dataset split sizes.}
\small
\resizebox{\textwidth}{!}{%
\begin{tabular}{lp{7.7cm}rrrr}
\toprule
\textbf{Dataset} & \textbf{Description} & \textbf{Classes} & \textbf{Train size} & \textbf{Valid size} & \textbf{Test size} \\
\midrule
IMDB~\cite{maas2011learning}                   & Sentiment classification of movie reviews. & 2  & 20000 & 5000  & 25000 \\
SST-5~\cite{socher-etal-2013-recursive}        & SC of movie-review sentences. & 5  & 8544  & 1101  & 2210 \\
20Newsgroups~\cite{LANG1995331}                & Topic classification for newsgroup posts. & 20 & 9051  & 2263  & 7532 \\
ToxiGen~\cite{hartvigsen-etal-2022-toxigen}    & Detection of toxicity and hate speech targeting groups detection. & 2  & 7168  & 1792  & 940 \\
CoLA~\cite{warstadt-etal-2019-neural}          & Grammatical acceptability for English sentences. & 2  & 6840  & 1711  & 1043 \\
Yelp~\cite{asghar2016yelp}                     & SC of Yelp user reviews (3-way variant). & 3  & 27300 & 11700 & 30000 \\
GoEmotions~\cite{demszky2020goemotions}        & Emotion classification of Reddit comments. & 27 & 5000  & 2000  & 2000 \\
CoNLL-2003~\cite{sang2003introduction}         & English newswire named entity recognition (NER). & 4  & 14041 & 3250  & 3453 \\
WNUT-2017~\cite{derczynski2017results}         & Emerging and rare NER on noisy social media text. & 6  & 3394  & 1009  & 1287 \\
\bottomrule
\end{tabular}
}
\label{table:datasets_sizes}
\end{table*}

\paragraph{Models.} We evaluate ALIEN on five pre-trained models: two smaller encoders, RoBERTa-base~\cite{liu2019roberta} (125M parameters; trained with Masked Language Modeling) and
ELECTRA-base~\cite{clark2020electric} (110M; Replaced Token Detection), as well as three decoder LLMs, Qwen2.5-7B~\cite{qwen2}, LLaMA-2-7B~\cite{touvron2023llama2openfoundation}, and Qwen3-1.7B~\cite{yang2025qwen3technicalreport}. For NER tasks we evaluate two pre-trained models:
distilbert-NER~\cite{sanh2019distilbert} for CoNLL-2003 and
bertweet-large-wnut2017~\cite{nguyen2020bertweet} for WNUT-2017.


\paragraph{Technical details.}  

All models employed parameter-efficient fine-tuning through LoRA, applied to attention mechanisms ($W_q$, $W_k$, $W_v$), with $\alpha = 16$, rank $= 8$, dropout $= 0.05$.
Optimization uses the AdamW optimizer with learning rate of $5\times10^{-4}$, $0.1$ weight decay, a batch size of $64$, and a linear LR schedule that ends at $0$ with a warm-up of $0.1$ steps.

\begin{table}[h]\scriptsize\setlength{\tabcolsep}{2.pt}
\caption{Mean inference time (milliseconds) for different uncertainty estimation methods evaluated on the same batch. ALIEN introduces minimal additional overhead compared to probability-based baselines while remaining significantly faster than distance-based methods.}
\begin{tabular}{lrrrrr}
\toprule
Method & Electra & Roberta & LLaMA & Qwen & Qwen3-1.7B \\
\midrule
MD & 61.32 $\pm$ 4.92 & 56.57 $\pm$ 0.17 & 7524.40 $\pm$ 10.29 & 5345.71 $\pm$ 9.96 & 1369.84 $\pm$ 8.82 \\
MDR & 103.72 $\pm$ 3.25 & 101.64 $\pm$ 0.17 & 12728.23 $\pm$ 16.50 & 8866.12 $\pm$ 14.48 & 2134.15 $\pm$ 4.21 \\
MDM & 45.18 $\pm$ 0.23 & 44.65 $\pm$ 0.33 & 5183.80 $\pm$ 9.63 & 3531.08 $\pm$ 13.75 & 769.81 $\pm$ 1.63 \\
RDE & 32.77 $\pm$ 0.00 & 27.51 $\pm$ 0.00 & 28.75 $\pm$ 0.00 & 28.24 $\pm$ 0.00 & 28.32 $\pm$ 0.00 \\
Linear probe & 0.03 $\pm$ 0.00 & 0.03 $\pm$ 0.00 & 0.03 $\pm$ 0.00 & 0.03 $\pm$ 0.00 & 0.03 $\pm$ 0.00 \\
Attention pooling & 0.39 $\pm$ 0.00 & 0.38 $\pm$ 0.00 & 4.90 $\pm$ 0.04 & 4.13 $\pm$ 0.03 & 1.98 $\pm$ 0.02 \\
Entropy & 0.03 $\pm$ 0.00 & 0.03 $\pm$ 0.00 & 0.03 $\pm$ 0.00 & 0.03 $\pm$ 0.00 & 0.03 $\pm$ 0.00 \\
\textbf{ALIEN (Ours)} & 0.36 $\pm$ 0.00 & 0.35 $\pm$ 0.00 & 0.10 $\pm$ 0.00 & 0.09 $\pm$ 0.00 & 0.08 $\pm$ 0.00 \\
\bottomrule
\end{tabular}
\label{tab:inference_speed_news}
\end{table}



We consider the hyperparameter grids to select from for probing and ALIEN methods.
\emph{ALIEN:} output regularization $\alpha \in \{0.01, 0.1, 1.0\}$,
L2-SP $\beta \in \{0.01, 0.1,$ $ 1.0\}$,
learning rate $\in \{4\times10^{-4}, 1\times10^{-4}, 1\times10^{-5}\}$,
20 training epochs.
\emph{Probes:} linear and attention-pooling with learning rate
$\in \{1\times10^{-3}, 4\times10^{-4}, 1\times10^{-4}\}$ for 20 epochs. Experiments were conducted on a server with 128 CPU Cores of 2 TB and 8 NVIDIA A100.
\subsection{Baselines}
\label{sec:baselines}
\paragraph{Probing methods} extend beyond representational analysis to sequence classification. A probe is an auxiliary classifier applied to intermediate hidden states to test whether internal representations encode information consistent with the conditioning prompt. We consider two probes: a linear probe and an attention-pooling probe. For each model, we attach probes at three depths: begin ($\approx$25\% of layers), middle ($\approx$50\%), and last (final layer).

\emph{Linear Probing} A linear probe classifies a single hidden state vector from a chosen sublayer~\cite{kadavath2022language}. 
We assign $y_i \in \{0, 1\}$ to indicate whether a sequence of tokens $\vecX_i$ belongs to a wrongly classified sample or not. 
The prediction is given by a logistic regression over $\mathbf{h}_i \in \mathbb{R}^n$, the representation of the last token at a selected layer.

\emph{Attention-Pooling Probing.} 
To include contextual information, we use an attention-pooling variant~\cite{sky2024androids} that aggregates prior hidden states:
\[
p(y_i = 1 \mid \vecX_i) = \sigma(\mathbf{w}^\top \bar{\mathbf{h}}_i),
\]
where for a learnable query vector $\mathbf{q} \in \mathbb{R}^n$:
\[
\bar{\mathbf{h}}_i = \sum_{j=1}^{i} \alpha_{i,j} \mathbf{h}_j, 
\quad
\alpha_{i,j} = \frac{\exp(\mathbf{q}^\top \mathbf{h}_j)}{\sum_{k=1}^{i} \exp(\mathbf{q}^\top \mathbf{h}_k)}.
\]

\paragraph{Distance-based UE methods.} Here, we describe the implementation as well as extensions of MD --- a Mahalanobis distance-based uncertainty estimation.

The MD method \cite{NEURIPS2018_abdeb6f5} models each class as a Gaussian distribution characterized by class centroids $\boldsymbol{\mu}_c = \mathbb{E}[h(\vecX) | y = c]$, 
a shared covariance matrix $\boldsymbol{\Sigma} = \mathbb{E}[(
\vecH - \boldsymbol{\mu}_c) (\vecH - \boldsymbol{\mu}_c)^T]$.

The uncertainty score for an input $\mathbf{x}$ is given by the minimum Mahalanobis distance to any class centroid:
\begin{equation}\label{eq:md}
    U_{\text{MD}}(\vecX) = \min_{c \in \mathcal{C}} (\vecH - \boldsymbol{\mu}_c)^T \boldsymbol{\Sigma}^{-1} (\vecH - \boldsymbol{\mu}_c).
\end{equation} where $\vecH$ is a representation from the penultimate layer.

Two methods extend MD to make it more robust are the \emph{Relative Mahalanobis Distance (MDR)}~\cite{ren2021simple} and the \emph{Robust Distance Estimation (RDE)}~\cite{yoo-etal-2022-detection}. These modifications enhance robustness to outliers while preserving the discriminative power of the original representations. \emph{(MDM) Marginal Mahalanobis} distance measures how far a feature vector $\vecH$
is from the class-agnostic (marginal) background distribution of training features.

We note that the MD method is very close to a linear layer on top of the representations, with appropriate scaling, followed by a soft minimum.
The outputs of this layer are similar in number to the outputs.
Thus, with an appropriate training procedure, we should be able to model epistemic uncertainty using a similar linear head - a goal that we want to achieve with our ALIEN head.

\subsection{Results for Text Classification}

The main results are presented in Table~\ref{tab:final_results_all} and Table~\ref{tab:final_results_all_rejection_curve} for missclassification ROC-AUC and areas for risk coverage curves, while the underlying task accuracies of the fine-tuned models are reported in Table~\ref{tab:accuracy_results}.
ALIEN achieves the highest ROC--AUC in 32 out of 35 dataset--model combinations, for both encoder and decoder-based models.
Similar superior performance is observed for rejection curves.
Entropy is the second-best baseline, despite numerous more complex methods being proposed.
We hypothesize that this outcome stems from the limited size of the fine-tuning datasets, which may prevent more complex uncertainty methods from realizing their full potential. 
In low-data settings, the model's own entropy remains strong, whereas methods that require additional training probes or fitting distance metrics overfit or lack sufficient data
volumes to excel. 
Our evaluation also considered calibration quality with Expected calibration error as the metric.
Due to the space constraints, we report only average ranks for ECE.
As Table~\ref{tab:ece_rank} reports, ALIEN attains the best average calibration across all compared methods. Also, ALIEN introduces minimal computational overhead and does not require storing additional activations, unlike attention-pooling probes; inference-time measurements are reported in Table~\ref{tab:inference_speed_news}. In practice, this overhead remains very small relative to the backbone size. For LLaMA-7B and Qwen-7B, ALIEN adds fewer than 100k parameters, corresponding to only 0.001\% of the full model. For Qwen3-1.7B, the additional module contains around 50k parameters, or 0.002\% of the model size. Even for the smaller encoder backbones, Electra and RoBERTa, the added parameters remain below 0.6\% of the total number of model parameters.

\begin{table*}[h]
\caption{ROC–AUC ↑ for misclassification detection across seven classification datasets and five model architectures. Results are reported as mean ± standard deviation over 20 bootstrap samples. ALIEN achieves the best or competitive performance in the majority of dataset–model combinations.}
\tiny\setlength{\tabcolsep}{3.0 pt}
\makebox[\textwidth][c]{
\begin{tabular}{lccccccccccccc}
\toprule
Dataset & \multicolumn{4}{c}{Mahalanobis distance} & \multicolumn{3}{c}{Linear probing} & \multicolumn{3}{c}{Attention pooling} & \multicolumn{3}{c}{Probability based} \\
 & MD & MDR & MDM & RDE & Last & Mid & Begin & Last & Mid & Begin & SR & Entropy & \textbf{ALIEN (Ours)} \\
\midrule
\multicolumn{14}{c}{\cellcolor[gray]{0.9} LLaMA2 7B} \\
Cola & 62.2± 1.4 & 69.9± 1.1 & 62.2± 1.4 & 68.0± 1.2 & 72.0± 1.0 & 60.6± 1.2 & 58.1± 1.1 & 61.5± 1.1 & 51.4± 1.2 & 60.3± 1.1 & 76.9± 1.0 & \underline{77.0± 0.9} & \textbf{78.0± 0.9} \\
GEmot & 49.5± 0.2 & 50.0± 0.2 & 49.5± 0.2 & 50.6± 0.2 & \underline{64.7± 0.4} & 64.4± 0.5 & 63.2± 0.3 & 61.6± 0.4 & 63.1± 0.4 & 63.0± 0.4 & 61.4± 0.2 & 59.2± 0.2 & \textbf{66.4± 0.2} \\
IMDB & 65.4± 0.3 & 75.8± 0.3 & 65.4± 0.3 & 62.9± 0.3 & 79.2± 0.2 & 78.6± 0.2 & 77.3± 0.2 & 84.7± 0.2 & 84.9± 0.2 & \underline{85.1± 0.1} & 83.5± 0.2 & 83.5± 0.2 & \textbf{88.7± 0.1} \\
News & 57.1± 0.3 & 80.1± 0.3 & 56.7± 0.3 & 46.6± 0.3 & 76.2± 0.3 & 69.4± 0.5 & 61.0± 0.5 & 78.7± 0.4 & 75.1± 0.4 & 67.5± 0.4 & 85.7± 0.2 & \underline{85.8± 0.2} & \textbf{88.8± 0.3} \\
SST5 & 51.9± 0.6 & 52.8± 0.6 & 51.9± 0.6 & 52.2± 0.7 & 56.2± 0.6 & 56.5± 0.6 & 56.5± 0.8 & 53.1± 0.5 & 52.1± 0.5 & 54.9± 0.5 & \underline{62.8± 0.6} & 60.0± 0.5 & \textbf{63.5± 0.6} \\
Tox & 53.0± 1.1 & 63.9± 1.1 & 53.0± 1.1 & 55.2± 1.5 & 66.7± 0.8 & 60.8± 1.2 & 56.2± 1.4 & 62.0± 0.9 & 43.0± 1.7 & 59.9± 1.4 & \underline{77.9± 1.1} & \underline{77.9± 1.1} & \textbf{81.6± 1.1} \\
YELP & 61.0± 0.2 & 79.9± 0.2 & 61.0± 0.2 & 62.6± 0.2 & 79.3± 0.2 & 77.4± 0.2 & 76.7± 0.2 & 80.5± 0.2 & 81.0± 0.2 & 81.9± 0.2 & \underline{83.8± 0.1} & \underline{83.8± 0.1} & \textbf{85.9± 0.1} \\
\midrule
\multicolumn{14}{c}{\cellcolor[gray]{0.9} Qwen2.5 7B} \\
Cola & 55.9± 0.8 & 59.7± 1.3 & 55.9± 0.8 & 51.3± 0.9 & 76.9± 0.9 & 69.0± 1.0 & 67.7± 1.3 & 75.6± 0.9 & 66.2± 1.0 & 64.3± 1.0 & \underline{78.9± 0.9} & \underline{78.9± 0.9} & \textbf{81.6± 0.9} \\
GEmot & 50.2± 0.3 & 50.4± 0.3 & 50.2± 0.3 & 48.3± 0.3 & 61.3± 0.3 & 60.7± 0.4 & 58.7± 0.4 & 54.9± 0.5 & 53.8± 0.4 & 55.7± 0.5 & \underline{66.1± 0.2} & 65.0± 0.2 & \textbf{68.2± 0.2} \\
IMDB & 65.1± 0.3 & 78.5± 0.2 & 65.1± 0.3 & 62.5± 0.3 & 83.3± 0.2 & 83.0± 0.2 & 80.9± 0.2 & 83.4± 0.2 & 84.1± 0.2 & \underline{84.7± 0.1} & 82.1± 0.2 & 82.1± 0.2 & \textbf{89.1± 0.2} \\
News & 53.3± 0.6 & 81.2± 0.2 & 52.8± 0.6 & 51.2± 0.5 & 71.3± 0.4 & 61.1± 0.4 & 57.7± 0.4 & 71.3± 0.3 & 68.8± 0.4 & 56.9± 0.4 & 84.9± 0.2 & \underline{85.1± 0.2} & \textbf{87.8± 0.2} \\
SST5 & 52.9± 0.7 & 54.0± 0.6 & 52.9± 0.7 & 50.9± 0.5 & 54.3± 0.7 & 54.1± 0.6 & 50.4± 0.8 & 51.1± 0.6 & 52.5± 0.6 & 49.4± 0.6 & \underline{60.9± 0.7} & 58.7± 0.8 & \textbf{62.7± 0.7} \\
Tox & 64.2± 1.2 & 65.7± 1.3 & 64.1± 1.2 & 46.2± 1.1 & 57.3± 0.9 & 53.7± 1.1 & 50.8± 1.4 & 55.7± 1.0 & 50.0± 1.3 & 51.3± 1.0 & 74.9± 1.0 & \textbf{76.4± 0.9} & \textbf{76.4± 0.9} \\
YELP & 62.5± 0.3 & 83.2± 0.2 & 62.5± 0.3 & 51.7± 0.3 & 82.2± 0.1 & 80.6± 0.2 & 74.8± 0.2 & 83.7± 0.1 & 83.0± 0.1 & 80.8± 0.1 & \underline{85.8± 0.2} & 85.7± 0.2 & \textbf{88.5± 0.1} \\
\midrule
\multicolumn{14}{c}{\cellcolor[gray]{0.9} Qwen3 1.7B} \\
Cola & 50.9± 1.0 & 64.3± 1.6 & 50.6± 1.0 & 47.0± 1.1 & 63.8± 1.2 & 56.3± 1.2 & 62.8± 1.0 & 58.0± 1.0 & 56.0± 1.4 & 54.9± 1.1 & 73.5± 0.9 & \textbf{73.8± 1.0} & \underline{73.7± 0.9} \\
GEmot & 48.5± 0.5 & 52.5± 0.4 & 48.5± 0.5 & 50.0± 0.6 & 60.9± 0.4 & 55.6± 0.3 & 56.9± 0.4 & 57.6± 0.4 & 53.7± 0.3 & 56.6± 0.3 & \underline{66.9± 0.4} & 65.9± 0.4 & \textbf{68.0± 0.4} \\
IMDB & 61.1± 0.2 & 82.4± 0.2 & 61.0± 0.2 & 61.1± 0.2 & 83.1± 0.2 & 67.5± 0.2 & 61.5± 0.3 & \underline{85.6± 0.2} & 80.4± 0.2 & 75.1± 0.2 & 85.2± 0.2 & 85.2± 0.2 & \textbf{87.2± 0.2} \\
News & 64.0± 0.4 & 84.3± 0.3 & 63.1± 0.4 & 57.3± 0.4 & 72.4± 0.3 & 57.6± 0.4 & 56.6± 0.4 & 73.5± 0.2 & 67.0± 0.4 & 64.4± 0.3 & \underline{85.6± 0.2} & \underline{85.6± 0.2} & \textbf{86.9± 0.3} \\
SST5 & 49.6± 0.8 & 57.1± 0.6 & 49.6± 0.8 & 48.6± 0.7 & 56.2± 0.6 & 55.8± 0.7 & 53.0± 0.4 & 55.0± 0.7 & 53.1± 0.6 & 49.6± 0.6 & \underline{61.0± 0.6} & 59.8± 0.6 & \textbf{63.3± 0.5} \\
Tox & 53.8± 1.1 & 70.4± 1.5 & 53.6± 1.1 & 47.6± 1.2 & 51.2± 1.2 & 56.2± 1.1 & 57.5± 1.2 & 56.1± 1.3 & 48.7± 1.1 & 53.0± 1.2 & 75.6± 0.8 & \underline{76.4± 0.8} & \textbf{77.3± 0.9} \\
YELP & 61.1± 0.2 & 82.4± 0.2 & 61.1± 0.2 & 61.2± 0.2 & 81.5± 0.1 & 67.1± 0.2 & 64.1± 0.2 & 82.8± 0.1 & 75.5± 0.2 & 70.2± 0.2 & \underline{84.7± 0.1} & 84.6± 0.1 & \textbf{86.7± 0.1} \\
\midrule
\multicolumn{14}{c}{\cellcolor[gray]{0.9} Electra} \\
Cola & \textbf{79.0± 1.0} & 65.6± 1.7 & \textbf{79.0± 1.0} & 78.7± 1.0 & 67.5± 1.3 & 62.3± 1.3 & 62.6± 1.6 & 66.2± 1.2 & 58.3± 1.3 & 51.9± 1.4 & 77.3± 0.8 & 77.3± 0.8 & 78.8± 0.7 \\
GEmot & 56.6± 0.2 & 55.1± 0.2 & 56.5± 0.2 & 55.5± 0.2 & 64.4± 0.2 & 60.3± 0.2 & 59.1± 0.2 & 64.4± 0.2 & 62.8± 0.2 & 62.2± 0.2 & \underline{66.4± 0.2} & 64.9± 0.2 & \textbf{68.1± 0.2} \\
IMDB & 83.5± 0.2 & 83.4± 0.2 & 83.5± 0.2 & 83.9± 0.2 & 85.2± 0.2 & 69.8± 0.3 & 69.2± 0.3 & \underline{86.0± 0.2} & 75.3± 0.2 & 72.6± 0.2 & 85.6± 0.2 & 85.6± 0.2 & \textbf{86.1± 0.1} \\
News & 74.1± 0.3 & \underline{84.4± 0.3} & 73.1± 0.3 & 65.6± 0.3 & 78.8± 0.4 & 74.2± 0.4 & 71.9± 0.4 & 78.4± 0.3 & 76.0± 0.4 & 75.4± 0.4 & 84.0± 0.2 & 84.3± 0.2 & \textbf{85.7± 0.1} \\
SST5 & 56.5± 0.5 & 57.3± 0.6 & 56.5± 0.5 & 57.2± 0.5 & 56.0± 0.5 & 51.3± 0.5 & 51.3± 0.6 & 56.6± 0.6 & 54.1± 0.5 & 51.9± 0.7 & \underline{61.3± 0.5} & 61.1± 0.6 & \textbf{62.9± 0.5} \\
Tox & \textbf{76.3± 1.0} & 69.4± 1.2 & \underline{76.2± 1.0} & 75.6± 1.1 & 48.3± 1.2 & 50.2± 1.4 & 51.9± 1.2 & 59.3± 1.3 & 62.0± 1.1 & 60.0± 1.0 & 75.2± 1.2 & 75.7± 1.2 & 74.8± 1.2 \\
YELP & 74.0± 0.2 & 82.9± 0.2 & 73.9± 0.2 & 76.1± 0.2 & 81.0± 0.1 & 64.4± 0.2 & 63.7± 0.1 & 81.4± 0.1 & 69.5± 0.2 & 68.2± 0.2 & 84.4± 0.2 & \underline{84.5± 0.2} & \textbf{85.6± 0.1} \\
\midrule
\multicolumn{14}{c}{\cellcolor[gray]{0.9} Roberta} \\
Cola & 73.6± 0.8 & 67.0± 1.7 & 73.5± 0.8 & 75.1± 0.9 & 71.4± 0.8 & 68.1± 1.1 & 63.8± 1.1 & 67.7± 1.0 & 71.0± 1.1 & 60.5± 1.4 & \underline{79.3± 1.2} & \underline{79.3± 1.2} & \textbf{80.5± 0.9} \\
GEmot & 51.8± 0.2 & 53.6± 0.3 & 51.7± 0.2 & 52.0± 0.2 & 62.1± 0.2 & 58.4± 0.2 & 55.8± 0.2 & 62.5± 0.2 & 61.4± 0.3 & 60.5± 0.2 & \underline{64.9± 0.2} & 63.2± 0.2 & \textbf{66.7± 0.2} \\
IMDB & 83.6± 0.3 & 77.9± 0.2 & 83.6± 0.3 & \underline{83.9± 0.2} & 83.5± 0.2 & 70.1± 0.2 & 65.3± 0.2 & 83.2± 0.2 & 77.6± 0.2 & 71.9± 0.3 & 81.8± 0.3 & 83.0± 0.2 & \textbf{85.2± 0.2} \\
News & 79.5± 0.3 & 84.4± 0.2 & 79.0± 0.3 & 76.6± 0.3 & 76.5± 0.3 & 68.9± 0.3 & 59.5± 0.3 & 76.7± 0.3 & 71.2± 0.3 & 60.8± 0.3 & 84.6± 0.3 & \underline{84.9± 0.3} & \textbf{86.0± 0.3} \\
SST5 & 56.3± 0.7 & 59.5± 0.5 & 56.2± 0.7 & 57.0± 0.7 & 58.6± 0.8 & 55.6± 0.6 & 52.7± 0.7 & 57.0± 0.8 & 56.9± 0.6 & 55.6± 0.6 & \underline{61.1± 0.5} & 60.4± 0.5 & \textbf{63.9± 0.6} \\
Tox & 57.7± 1.0 & 68.5± 1.0 & 57.6± 1.0 & 57.6± 1.1 & 71.0± 0.8 & 57.9± 1.0 & 53.4± 1.0 & 74.2± 0.8 & 63.8± 0.9 & 64.4± 1.1 & \underline{74.9± 1.0} & \underline{74.9± 1.0} & \textbf{75.0± 1.1} \\
YELP & 79.5± 0.2 & 80.7± 0.1 & 79.4± 0.2 & 79.0± 0.2 & 73.2± 0.2 & 61.5± 0.2 & 59.3± 0.2 & 73.8± 0.2 & 66.9± 0.2 & 64.2± 0.3 & 83.0± 0.2 & \underline{83.3± 0.2} & \textbf{84.2± 0.1} \\
\midrule
Rank & 9.1 & 6.7 & 9.6 & 9.9 & 6.0 & 8.7 & 9.9 & 6.1 & 8.4 & 8.9 & 3.3 & \underline{3.2} & \textbf{1.2} \\
\bottomrule
\end{tabular}
}
\label{tab:final_results_all}
\end{table*}

\begin{table*}[h]
\caption{Accuracy ↑ of classification task (mean $\pm$ std over 20 bootstraps) for each model and dataset.}
\scriptsize\setlength{\tabcolsep}{3. pt}
\begin{tabular}{lrrrrrrr}
\toprule
 & Cola & GEmot & IMDB & News & SST5 & Toxigen & YELP \\
\midrule
Electra & 85.97± 0.45 & 35.35± 0.18 & 90.37± 0.1 & 71.92± 0.29 & 54.96± 0.38 & 81.2± 0.61 & 86.84± 0.11 \\
LLaMA2 7B & 84.86± 0.62 & 27.7± 0.15 & 89.73± 0.12 & 74.75± 0.33 & 59.36± 0.61 & 84.36± 0.64 & 88.12± 0.09 \\
Qwen2.5 7B & 84.04± 0.65 & 37.3± 0.16 & 90.17± 0.11 & 79.49± 0.22 & 58.4± 0.32 & 85.84± 0.45 & 88.78± 0.08 \\
Qwen3 1.7B & 82.19± 0.64 & 39.45± 0.29 & 91.06± 0.09 & 81.11± 0.18 & 56.29± 0.56 & 81.2± 0.52 & 87.24± 0.11 \\
Roberta & 83.02± 0.51 & 39.11± 0.14 & 89.8± 0.08 & 74.65± 0.21 & 55.1± 0.57 & 78.98± 0.7 & 85.7± 0.07 \\
\bottomrule
\end{tabular}
\label{tab:accuracy_results}
\end{table*}

\begin{table}[h!]
\caption{Average Expected Calibration Error (ECE) ↓ across datasets and models. Lower values indicate better calibration. ALIEN achieves the best average calibration performance among the compared uncertainty estimation methods.}
\scriptsize
\setlength{\tabcolsep}{3.0 pt}
\makebox[\textwidth][c]{
\begin{tabular}{lccccccccccccc}
\toprule
 \multicolumn{4}{c}{Mahalanobis distance} & \multicolumn{3}{c}{Linear probing} & \multicolumn{3}{c}{Attention pooling} & \multicolumn{3}{c}{Probability based} \\
 MD & MDR & MDM & RDE & Last & Mid & Begin & Last & Mid & Begin & SR & Entropy & ALIEN (Ours) \\
\midrule
19.2$\pm$19.4 & 14.9$\pm$15.8 & 19.3$\pm$19.4 & 20.7$\pm$18.1 & 9.9$\pm$5.8 & 10.8$\pm$6.6 & 10.5$\pm$7.2 & 10.3$\pm$5.3 & 10.5$\pm$5.6 & 12.2$\pm$10.6 & \underline{9.4$\pm$5.1} & 12.6$\pm$5.0 & \textbf{8.4$\pm$4.7} \\
\bottomrule
\end{tabular}
}

\label{tab:ece_rank}
\end{table}
\begin{table*}[t]
\caption{Area under the risk–coverage curve ↓ for misclassification detection across classification datasets and model architectures. Lower values indicate better performance. ALIEN consistently produces lower risk–coverage AUC.}
\tiny\setlength{\tabcolsep}{3. pt}
\makebox[\textwidth][c]{
\begin{tabular}{lrrrrrrrrrrrrrr}
\toprule
Dataset & \multicolumn{4}{c}{Mahalanobis distance} & \multicolumn{3}{c}{Linear probing} & \multicolumn{3}{c}{Attention pooling} & \multicolumn{4}{c}{Probability based} \\
 & MD & MDR & MDM & RDE & Last & Mid & Begin & Last & Mid & Begin & SR & Entropy & \textbf{ALIEN (Ours)} & Oracle \\
\midrule
\multicolumn{15}{c}{\cellcolor[gray]{0.9} LLaMA2 7B} \\
Cola & 10.3± 0.5 & 9.3± 0.7 & 10.4± 0.5 & 8.4± 0.4 & 7.5± 0.4 & 13.6± 0.6 & 12.6± 0.5 & 11.2± 0.5 & 15.7± 0.6 & 11.2± 0.5 & 6.8± 0.5 & \underline{6.5± 0.4} & \textbf{6.2± 0.4} & 1.2± 0.1 \\
GEmot & 72.7± 0.2 & 72.7± 0.2 & 72.7± 0.2 & 72.2± 0.2 & \underline{62.0± 0.5} & 62.1± 0.5 & 63.9± 0.6 & 64.8± 0.4 & 64.0± 0.6 & 63.8± 0.6 & 64.6± 0.2 & 65.9± 0.2 & \textbf{60.9± 0.3} & 36.7± 0.2 \\
IMDB & 6.2± 0.1 & 4.4± 0.1 & 6.2± 0.1 & 6.5± 0.1 & 3.3± 0.0 & 3.3± 0.0 & 3.6± 0.0 & 2.5± 0.0 & \underline{2.4± 0.0} & \underline{2.4± 0.0} & 2.8± 0.0 & 2.8± 0.0 & \textbf{2.0± 0.0} & 0.6± 0.0 \\
News & 21.1± 0.3 & 10.2± 0.2 & 21.4± 0.3 & 24.7± 0.3 & 11.4± 0.2 & 14.7± 0.3 & 19.4± 0.4 & 10.5± 0.2 & 12.1± 0.2 & 16.0± 0.3 & 8.0± 0.2 & \underline{7.9± 0.2} & \textbf{7.0± 0.1} & 3.5± 0.1 \\
SST5 & 38.8± 0.7 & 37.8± 0.5 & 38.8± 0.7 & 39.0± 0.7 & 36.7± 0.8 & 35.8± 0.6 & 35.2± 0.7 & 37.6± 0.7 & 37.0± 0.7 & 35.9± 0.7 & \underline{31.1± 0.5} & 32.5± 0.5 & \textbf{30.4± 0.6} & 9.8± 0.3 \\
Tox & 13.2± 0.9 & 10.0± 0.6 & 13.2± 0.9 & 13.2± 0.8 & 9.2± 0.6 & 12.2± 0.7 & 14.4± 0.7 & 10.9± 0.8 & 17.7± 0.7 & 11.6± 0.6 & \underline{5.7± 0.4} & 5.9± 0.4 & \textbf{5.2± 0.5} & 1.2± 0.1 \\
YELP & 8.5± 0.1 & 4.6± 0.1 & 8.6± 0.1 & 7.6± 0.1 & 3.8± 0.0 & 4.2± 0.0 & 4.3± 0.0 & 3.7± 0.0 & 3.7± 0.0 & 3.4± 0.0 & \underline{3.3± 0.0} & \underline{3.3± 0.0} & \textbf{2.8± 0.0} & 0.7± 0.0 \\
\midrule
\multicolumn{15}{c}{\cellcolor[gray]{0.9} Qwen2.5 7B} \\
Cola & 14.5± 0.9 & 13.8± 0.6 & 14.4± 0.9 & 16.5± 0.9 & 6.0± 0.4 & 8.2± 0.6 & 8.5± 0.4 & 7.2± 0.5 & 9.7± 0.7 & 9.8± 0.6 & \underline{5.7± 0.3} & \underline{5.7± 0.3} & \textbf{5.3± 0.3} & 1.3± 0.1 \\
GEmot & 61.8± 0.3 & 62.2± 0.3 & 61.8± 0.2 & 63.0± 0.3 & 53.7± 0.5 & 53.3± 0.5 & 54.5± 0.4 & 58.5± 0.4 & 59.1± 0.4 & 57.1± 0.5 & \underline{49.8± 0.2} & 50.3± 0.2 & \textbf{48.3± 0.2} & 25.9± 0.2 \\
IMDB & 5.9± 0.1 & 3.9± 0.1 & 5.9± 0.1 & 6.3± 0.1 & 2.6± 0.0 & 2.5± 0.0 & 2.8± 0.0 & 2.6± 0.0 & 2.5± 0.0 & \underline{2.4± 0.0} & 3.1± 0.1 & 3.1± 0.1 & \textbf{1.9± 0.0} & 0.5± 0.0 \\
News & 19.0± 0.3 & 7.2± 0.1 & 19.4± 0.3 & 19.1± 0.3 & 10.7± 0.1 & 15.1± 0.3 & 16.6± 0.2 & 10.5± 0.1 & 11.9± 0.2 & 18.1± 0.3 & 6.1± 0.1 & \underline{6.0± 0.1} & \textbf{5.3± 0.1} & 2.2± 0.1 \\
SST5 & 39.8± 0.5 & 37.5± 0.8 & 39.8± 0.5 & 41.5± 0.6 & 39.5± 0.9 & 38.9± 0.7 & 40.7± 0.9 & 41.7± 0.6 & 40.3± 0.6 & 41.9± 0.6 & \underline{32.4± 0.6} & 33.2± 0.7 & \textbf{31.1± 0.8} & 10.1± 0.2 \\
Tox & 9.0± 0.5 & 9.4± 0.7 & 9.1± 0.5 & 16.0± 0.8 & 10.2± 0.6 & 12.3± 0.7 & 13.4± 0.7 & 11.6± 0.8 & 13.5± 0.7 & 13.2± 0.7 & 6.7± 0.8 & \textbf{5.5± 0.5} & \underline{5.8± 0.7} & 1.0± 0.1 \\
YELP & 7.7± 0.1 & 3.6± 0.1 & 7.7± 0.1 & 11.6± 0.1 & 3.0± 0.0 & 3.4± 0.0 & 4.4± 0.1 & 2.8± 0.0 & 3.0± 0.0 & 3.4± 0.0 & \underline{2.6± 0.0} & \underline{2.6± 0.0} & \textbf{2.2± 0.0} & 0.7± 0.0 \\
\midrule
\multicolumn{15}{c}{\cellcolor[gray]{0.9} Qwen3 1.7B} \\
Cola & 16.0± 0.7 & 13.8± 0.7 & 16.1± 0.7 & 18.2± 0.8 & 11.8± 0.7 & 15.4± 0.9 & 11.6± 0.6 & 14.0± 0.7 & 16.6± 0.8 & 16.2± 0.9 & 9.2± 0.6 & \textbf{8.5± 0.4} & \underline{9.0± 0.6} & 1.7± 0.1 \\
GEmot & 61.4± 0.3 & 58.9± 0.4 & 61.5± 0.3 & 59.5± 0.3 & 50.4± 0.6 & 55.0± 0.5 & 54.9± 0.4 & 55.6± 0.5 & 56.9± 0.5 & 55.0± 0.5 & \underline{46.2± 0.6} & 46.7± 0.6 & \textbf{45.7± 0.6} & 23.8± 0.4 \\
IMDB & 9.4± 0.1 & 4.2± 0.1 & 9.5± 0.1 & 9.1± 0.1 & 2.3± 0.0 & 4.8± 0.1 & 6.1± 0.1 & \underline{2.0± 0.0} & 2.7± 0.1 & 3.6± 0.1 & 2.1± 0.0 & 2.1± 0.0 & \textbf{1.8± 0.0} & 0.4± 0.0 \\
News & 11.8± 0.2 & 5.7± 0.1 & 12.3± 0.2 & 13.4± 0.2 & 9.9± 0.2 & 16.2± 0.4 & 16.4± 0.4 & 9.8± 0.2 & 12.0± 0.3 & 13.0± 0.3 & \underline{5.2± 0.1} & \underline{5.2± 0.1} & \textbf{4.9± 0.1} & 1.9± 0.0 \\
SST5 & 45.0± 0.8 & 37.6± 0.9 & 45.0± 0.8 & 45.3± 0.8 & 38.5± 0.8 & 40.1± 0.7 & 41.6± 0.7 & 39.2± 0.8 & 42.1± 0.7 & 44.6± 1.0 & \underline{34.6± 0.6} & 35.1± 0.7 & \textbf{32.8± 0.6} & 11.2± 0.3 \\
Tox & 16.5± 1.0 & 12.0± 0.7 & 16.6± 1.0 & 18.8± 0.9 & 16.2± 1.0 & 15.6± 0.8 & 15.3± 0.8 & 15.0± 1.0 & 19.1± 0.8 & 17.7± 0.7 & 9.2± 0.7 & \textbf{8.3± 0.3} & \underline{8.5± 0.6} & 1.9± 0.1 \\
YELP & 9.4± 0.1 & 4.2± 0.1 & 9.5± 0.1 & 9.1± 0.1 & 3.8± 0.0 & 7.2± 0.1 & 7.9± 0.1 & 3.5± 0.0 & 5.0± 0.1 & 6.4± 0.1 & \underline{3.2± 0.0} & \underline{3.2± 0.0} & \textbf{2.9± 0.0} & 0.9± 0.0 \\
\midrule
\multicolumn{15}{c}{\cellcolor[gray]{0.9} Electra} \\
Cola & \textbf{4.9± 0.4} & 10.7± 0.7 & \textbf{4.9± 0.4} & 5.3± 0.5 & 7.8± 0.5 & 9.4± 0.5 & 10.2± 0.6 & 8.3± 0.5 & 11.9± 0.7 & 14.2± 0.9 & 5.2± 0.3 & 5.2± 0.3 & \textbf{4.9± 0.3} & 1.0± 0.1 \\
GEmot & 60.1± 0.3 & 60.8± 0.2 & 60.2± 0.3 & 61.2± 0.3 & \textbf{49.6± 0.2} & 53.0± 0.2 & 53.6± 0.2 & \underline{49.7± 0.2} & 50.8± 0.2 & 51.2± 0.2 & 52.8± 0.3 & 53.5± 0.3 & 50.8± 0.2 & 28.0± 0.2 \\
IMDB & 2.5± 0.0 & 3.0± 0.1 & 2.5± 0.0 & 2.4± 0.0 & 2.2± 0.0 & 5.0± 0.1 & 5.1± 0.1 & \textbf{2.1± 0.0} & 3.7± 0.1 & 4.3± 0.1 & 2.2± 0.0 & 2.2± 0.0 & \textbf{2.1± 0.0} & 0.5± 0.0 \\
News & 14.0± 0.2 & 10.6± 0.2 & 14.4± 0.3 & 17.0± 0.3 & 12.6± 0.2 & 14.9± 0.2 & 15.4± 0.3 & 12.7± 0.2 & 13.9± 0.3 & 14.0± 0.3 & 10.3± 0.2 & \underline{10.2± 0.2} & \textbf{9.6± 0.2} & 4.4± 0.1 \\
SST5 & 39.3± 0.7 & 38.4± 0.7 & 39.3± 0.7 & 38.4± 0.7 & 40.0± 0.6 & 43.0± 0.7 & 43.6± 0.8 & 38.8± 0.7 & 41.0± 0.7 & 42.8± 0.8 & 34.8± 0.7 & \underline{34.7± 0.8} & \textbf{33.7± 0.7} & 12.1± 0.3 \\
Tox & \textbf{7.9± 0.5} & 12.3± 0.7 & \textbf{7.9± 0.5} & 8.3± 0.5 & 21.3± 1.0 & 19.8± 0.7 & 19.1± 0.8 & 14.7± 0.9 & 14.0± 0.7 & 14.9± 0.7 & 9.1± 0.7 & 8.1± 0.4 & 8.3± 0.5 & 1.8± 0.1 \\
YELP & 5.7± 0.1 & 4.3± 0.1 & 5.7± 0.1 & 5.2± 0.1 & 4.0± 0.0 & 8.5± 0.1 & 8.6± 0.1 & 4.0± 0.0 & 6.6± 0.1 & 7.0± 0.1 & \underline{3.5± 0.1} & \underline{3.5± 0.1} & \textbf{3.2± 0.1} & 0.9± 0.0 \\
\midrule
\multicolumn{15}{c}{\cellcolor[gray]{0.9} Roberta} \\
Cola & 8.2± 0.5 & 12.1± 0.6 & 8.2± 0.5 & 7.8± 0.5 & 7.8± 0.3 & 9.1± 0.5 & 11.2± 0.7 & 9.0± 0.3 & 8.5± 0.4 & 12.8± 0.5 & \underline{6.2± 0.3} & \underline{6.2± 0.3} & \textbf{6.0± 0.3} & 1.5± 0.1 \\
GEmot & 57.4± 0.2 & 56.7± 0.2 & 57.7± 0.2 & 57.2± 0.2 & 50.5± 0.2 & 52.7± 0.2 & 54.8± 0.3 & 49.7± 0.2 & 50.3± 0.2 & 51.2± 0.2 & \underline{48.8± 0.2} & 49.6± 0.2 & \textbf{47.0± 0.2} & 24.1± 0.1 \\
IMDB & 2.8± 0.1 & 4.2± 0.1 & 2.8± 0.1 & 2.7± 0.1 & \underline{2.6± 0.0} & 5.3± 0.1 & 6.4± 0.1 & 2.7± 0.0 & 3.6± 0.1 & 4.8± 0.1 & 3.5± 0.1 & 3.0± 0.1 & \textbf{2.5± 0.0} & 0.5± 0.0 \\
News & 9.9± 0.2 & 8.6± 0.2 & 10.1± 0.2 & 11.0± 0.2 & 11.8± 0.2 & 15.0± 0.3 & 20.3± 0.3 & 11.5± 0.2 & 14.2± 0.2 & 19.5± 0.4 & \underline{8.2± 0.2} & \underline{8.2± 0.2} & \textbf{7.9± 0.2} & 3.5± 0.1 \\
SST5 & 39.7± 0.6 & 37.3± 0.7 & 39.8± 0.6 & 38.9± 0.6 & 38.3± 0.7 & 40.2± 0.7 & 42.8± 0.8 & 38.6± 0.6 & 39.0± 0.7 & 39.6± 0.6 & \underline{34.9± 0.6} & 35.7± 0.5 & \textbf{33.4± 0.6} & 11.9± 0.3 \\
Tox & 17.8± 0.8 & 13.5± 0.7 & 17.9± 0.8 & 18.0± 0.8 & 10.4± 0.3 & 15.8± 0.7 & 18.3± 0.8 & 9.7± 0.4 & 13.0± 0.5 & 13.1± 0.4 & \textbf{9.5± 0.6} & \textbf{9.5± 0.6} & \textbf{9.5± 0.5} & 2.4± 0.2 \\
YELP & 4.8± 0.1 & 5.4± 0.1 & 4.8± 0.1 & 4.9± 0.1 & 6.6± 0.1 & 11.6± 0.1 & 12.2± 0.1 & 6.6± 0.1 & 8.7± 0.1 & 9.4± 0.1 & 4.3± 0.1 & \underline{4.1± 0.1} & \textbf{3.9± 0.1} & 1.1± 0.0 \\
\midrule
Rank & 9.0 & 7.5 & 9.5 & 10.0 & 5.7 & 8.6 & 9.8 & 6.0 & 8.3 & 8.7 & 3.4 & \underline{3.3} & \textbf{1.3} & \underline{\textbf{-}} \\
\bottomrule
\end{tabular}
}
\label{tab:final_results_all_rejection_curve}
\end{table*}

\subsection{Results for Named Entity Recognition}
\label{sec:ner_experiments}

\begin{table}
\caption{ROC–AUC ↑ for misclassification detection on named entity recognition benchmarks. ALIEN achieves the best overall performance across both datasets.}
\centering
\scriptsize
\begin{tabular}{lrrc}
\toprule
 & conll2003 & wnut2017 & Rank \\
\midrule
Linear Probe & 70.4$\pm$1.6 & 78.0$\pm$1.4 & 5.9 \\
Entropy & 86.5$\pm$0.9 & 69.1$\pm$1.1 & 4.7 \\
MDM & 86.5$\pm$0.9 & 79.6$\pm$1.6 & 4.2 \\
MDR & 85.9$\pm$1.1 & \underline{83.7$\pm$1.2} & 3.5 \\
MD & \underline{86.9$\pm$0.9} & 80.2$\pm$1.6 & \underline{2.9} \\
SR & 86.4$\pm$0.8 & 68.9$\pm$1.1 & 5.8 \\
\textbf{ALIEN (Ours)} & \textbf{89.6$\pm$1.0} & \textbf{91.1$\pm$0.8} & \textbf{1.0} \\
\bottomrule
\end{tabular}
\label{tab:ner_results}
\end{table}
While our primary evaluation focuses on sentence-level classification, we also
assess the generalizability of ALIEN to token-level sequence labeling tasks,
specifically Named Entity Recognition. The results are
presented in Table~\ref{tab:ner_results}.

ALIEN achieves the best average rank, outperforming all baselines on
both benchmarks. On CoNLL-2003, ALIEN reaches 89.6 ROC--AUC, and on
WNUT-2017—a noisier, harder domain—it reaches 91.1, demonstrating that the
approach transfers well beyond sentence classification to per-token uncertainty
estimation. Notably, for WNUT-2017, probability-based
baselines (SR, Entropy) degrade substantially, whereas ALIEN maintains strong performance.

\subsection{Capturing both types of uncertainty}

Our results indicate that ALIEN better captures different types of uncertainty than the raw-entropy baseline, achieving higher scores. 
We conducted an additional experiment to identify which part of uncertainty ALIEN captures better.
To do so, we consider ensemble estimates for aleatoric $H_{\mathrm{alea}}$ and epistemic $H_{\mathrm{epi}}$ uncertainties.

\label{sec:ensemble_uncertainty}

To decompose uncertainty in practice, we use an ensemble of $T$ independently trained models and treat their predictions as Monte Carlo samples from an approximate posterior~\cite{lakshminarayanan2017simple}. 
For each input $\vecX$, the $t$-th ensemble member produces a vector of predicted class probabilities
\[
\vecP_t = p(\mathbf{c} \mid \vecX, \theta_t),
\]
where $\theta_t$ denotes the parameters of the $t$-th model with $\mathbf{c}$ is the set of considered classes.  
We then estimate the ensemble predictive distribution by averaging the member-wise predictive probabilities:
\[
    \bar{\vecP}(y) = \frac{1}{T} \sum_{t=1}^{T} \vecP_t.
\]
We consider the three uncertainty scores:
\begin{align*}
 H_{\mathrm{total}} &= H(\bar{\vecP}),  
 H_{\mathrm{alea}}  = \frac{1}{T} \sum_{t = 1}^{T} H(\vecP_t), 
 H_{\mathrm{epi}}   = H_{\mathrm{total}} - H_{\mathrm{alea}}.
\end{align*}
The predictive entropy $H_{\mathrm{total}}$ is computed as the entropy of this averaged distribution, and reflects the total predictive uncertainty. 
The aleatoric component $H_{\mathrm{alea}}$ is estimated as the average entropy of the individual ensemble members, which aims to capture irreducible data ambiguity shared across models. Finally, the epistemic component $H_{\mathrm{epi}}$ is obtained as the difference between total and aleatoric uncertainty, following~\cite{houlsby2011bayesian},

\paragraph{Technical details}
In our experiments, we use an ensemble of size $T = 5$ with models trained in the same way, but with different initializations on top of a corresponding backbone.
The three quantities above are computed for each test example and then used as uncertainty measures when comparing the behavior of \textsc{ALIEN} and entropy-based single-model scores.

\begin{figure}[h!]
  \centering
  \includegraphics[width=0.95\textwidth]{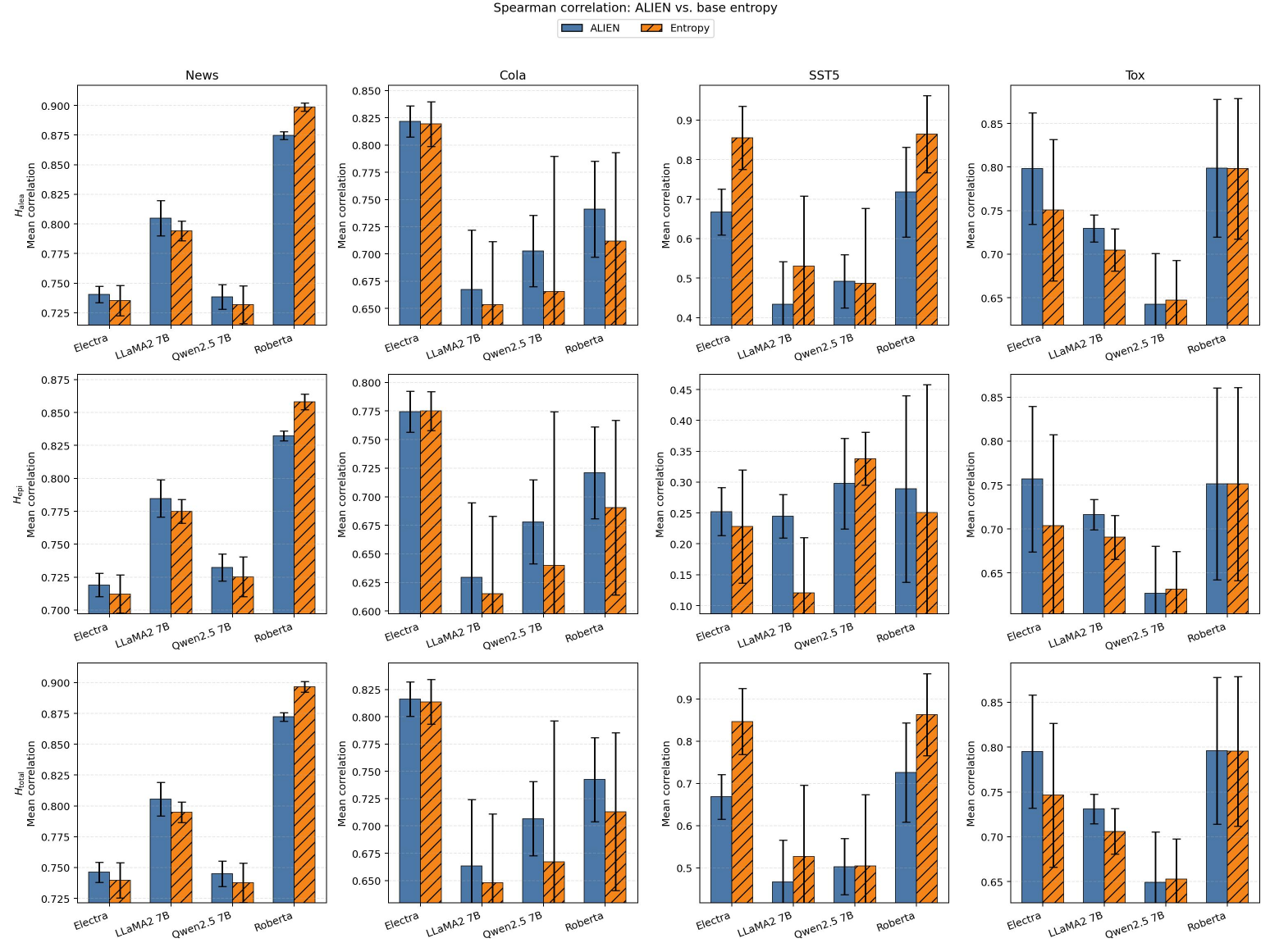}
  \caption{Spearman correlation between uncertainty estimates and ensemble-based uncertainty components across datasets. Each row corresponds to one uncertainty component ($H_{\mathrm{alea}}$, $H_{\mathrm{epi}}$, and $H_{\mathrm{total}}$), and each column corresponds to a dataset. Bars compare \textsc{ALIEN} against the base entropy across models. Higher values indicate stronger monotonic alignment with the corresponding uncertainty component. Best viewed when zoomed in.}
  \label{fig:alien_entropy_corr}
\end{figure}

Sample Spearman correlation coefficients for ALIEN and vanilla entropy with estimates of different uncertainty types are in Figure~\ref{fig:alien_entropy_corr}.
We see that ALIEN and Entropy have similar correlations with ensemble-based $H_{\mathrm{total}}$ and $H_{\mathrm{alea}}$ components.
For epistemic $H_{\mathrm{epi}}$, ALIEN provides better correlations.
Thus, by fine-tuning the entropy head, we can better capture epistemic uncertainty that was missed in the initial entropy score.

\subsection{Sensitivity Study}

\begin{table}[h]
\caption{Mean ROC–AUC ± standard deviation over 20 bootstrap samples for the ablation study on classification datasets. Results compare alternative head initializations and loss components. The full ALIEN model achieves the best average rank.}
\tiny\setlength{\tabcolsep}{3.0 pt}
\makebox[\textwidth][c]{
\begin{tabular}{lcccccc}
\toprule
Dataset & \shortstack{Trained CLS.\\BCE+L2SP} & \shortstack{Trained CLS.\\BCE+reg} & \shortstack{Trained CLS.\\BCE} & \shortstack{Rand CLS.\\BCE} & \shortstack{Rand linear.\\BCE} & \shortstack{\textbf{ALIEN (Ours)}} \\
\midrule
\multicolumn{7}{c}{\cellcolor[gray]{0.9} LLaMA2 7B} \\
Cola & 77.6± 1.2 & 77.5± 1.3 & \textbf{78.0± 0.9} & 60.5± 1.3 & 69.6± 1.0 & \textbf{78.0± 0.9} \\
GEmot & 65.2± 0.2 & \underline{65.8± 0.2} & 65.1± 0.2 & 63.1± 0.3 & 63.6± 0.2 & \textbf{66.4± 0.2} \\
IMDB & 84.6± 0.1 & \underline{84.7± 0.2} & 84.5± 0.2 & 59.4± 0.3 & 79.5± 0.2 & \textbf{88.7± 0.1} \\
News & 88.5± 0.2 & \underline{88.6± 0.2} & \underline{88.6± 0.2} & 74.2± 0.3 & 76.5± 0.3 & \textbf{88.8± 0.3} \\
SST5 & 62.2± 0.6 & \underline{63.0± 0.4} & 59.3± 0.5 & 53.8± 0.6 & 58.3± 0.9 & \textbf{63.5± 0.6} \\
Toxigen & \underline{81.0± 1.1} & 80.2± 0.9 & 80.6± 1.1 & 52.9± 1.4 & 65.7± 1.2 & \textbf{81.6± 1.1} \\
YELP & 85.7± 0.2 & \underline{85.8± 0.1} & 85.7± 0.2 & 79.2± 0.2 & 79.4± 0.2 & \textbf{85.9± 0.1} \\
\midrule
\multicolumn{7}{c}{\cellcolor[gray]{0.9} Qwen2.5 7B} \\
Cola & \underline{81.8± 0.9} & 81.6± 0.7 & \textbf{82.0± 1.0} & 79.7± 0.6 & 78.3± 1.1 & 81.6± 0.9 \\
GEmot & \textbf{68.4± 0.2} & 68.0± 0.2 & 68.0± 0.2 & 61.3± 0.2 & 60.5± 0.2 & \underline{68.2± 0.2} \\
IMDB & \underline{86.4± 0.2} & 85.0± 0.2 & 84.8± 0.2 & 78.8± 0.2 & 83.5± 0.2 & \textbf{89.1± 0.2} \\
News & \textbf{87.9± 0.2} & 87.8± 0.3 & \textbf{87.9± 0.2} & 72.9± 0.4 & 71.4± 0.4 & 87.8± 0.2 \\
SST5 & \textbf{63.8± 0.6} & 60.1± 0.6 & \underline{63.5± 0.6} & 51.4± 0.6 & 55.5± 0.4 & 62.7± 0.7 \\
Toxigen & \textbf{76.4± 1.3} & 76.3± 1.1 & 76.1± 1.0 & 56.0± 1.6 & 53.6± 0.9 & \textbf{76.4± 0.9} \\
YELP & \textbf{88.5± 0.1} & 88.4± 0.2 & \textbf{88.5± 0.1} & 82.2± 0.2 & 82.3± 0.1 & \textbf{88.5± 0.1} \\
\midrule
\multicolumn{7}{c}{\cellcolor[gray]{0.9} Qwen3 1.7B} \\
Cola & \textbf{73.8± 0.9} & \underline{73.7± 0.9} & 73.4± 0.8 & 71.7± 0.9 & 59.5± 1.5 & \underline{73.7± 0.9} \\
GEmot & \underline{67.6± 0.4} & 67.3± 0.3 & 67.5± 0.3 & 64.4± 0.4 & 62.2± 0.3 & \textbf{68.0± 0.4} \\
IMDB & \underline{86.2± 0.2} & 85.9± 0.2 & 85.5± 0.2 & 81.3± 0.2 & 83.2± 0.2 & \textbf{87.2± 0.2} \\
News & 86.7± 0.2 & \textbf{86.9± 0.2} & 86.8± 0.3 & 77.2± 0.4 & 73.2± 0.4 & \textbf{86.9± 0.3} \\
SST5 & \underline{64.3± 0.6} & 63.3± 0.5 & \textbf{64.6± 0.5} & 56.4± 0.3 & 59.3± 0.4 & 63.3± 0.5 \\
Toxigen & \underline{77.2± 1.0} & 76.9± 0.9 & 76.5± 1.1 & 45.4± 1.3 & 55.9± 1.3 & \textbf{77.3± 0.9} \\
YELP & 86.7± 0.1 & \textbf{86.8± 0.1} & \textbf{86.8± 0.1} & 81.5± 0.1 & 81.5± 0.2 & 86.7± 0.1 \\
\midrule
\multicolumn{7}{c}{\cellcolor[gray]{0.9} Electra} \\
Cola & \textbf{79.1± 0.9} & 79.0± 1.2 & 78.4± 0.8 & \textbf{79.1± 1.0} & 61.1± 1.2 & 78.8± 0.7 \\
GEmot & 67.8± 0.2 & 67.8± 0.2 & \underline{67.9± 0.2} & 65.9± 0.2 & 65.5± 0.2 & \textbf{68.1± 0.2} \\
IMDB & 85.7± 0.2 & 85.9± 0.2 & 85.9± 0.2 & \textbf{86.4± 0.2} & 85.2± 0.2 & \underline{86.1± 0.1} \\
News & 85.6± 0.2 & \textbf{86.3± 0.2} & \underline{86.2± 0.2} & 83.5± 0.3 & 77.7± 0.2 & 85.7± 0.1 \\
SST5 & 61.2± 0.6 & \textbf{62.9± 0.6} & 60.1± 0.8 & 58.6± 0.6 & 56.4± 0.5 & \textbf{62.9± 0.5} \\
Toxigen & 74.7± 0.8 & \underline{75.1± 1.3} & 74.8± 0.9 & 72.4± 1.1 & 50.2± 1.4 & \textbf{76.1± 0.8} \\
YELP & 85.4± 0.1 & \textbf{85.6± 0.1} & 85.5± 0.2 & 84.9± 0.1 & 80.4± 0.1 & \textbf{85.6± 0.1} \\
\midrule
\multicolumn{7}{c}{\cellcolor[gray]{0.9} Roberta} \\
Cola & 80.1± 0.7 & \textbf{80.5± 0.7} & 80.4± 0.9 & 68.3± 0.8 & 68.3± 0.8 & \textbf{80.5± 0.9} \\
GEmot & \textbf{66.7± 0.2} & \textbf{66.7± 0.2} & 66.4± 0.2 & 63.7± 0.2 & 62.8± 0.2 & \textbf{66.7± 0.2} \\
IMDB & 83.2± 0.3 & \underline{85.1± 0.2} & 82.8± 0.2 & 83.7± 0.2 & 83.6± 0.2 & \textbf{85.2± 0.2} \\
News & \textbf{86.1± 0.2} & 85.9± 0.2 & \textbf{86.1± 0.3} & 78.8± 0.3 & 76.0± 0.3 & 86.0± 0.3 \\
SST5 & \underline{63.8± 0.5} & 62.8± 0.6 & 62.8± 0.6 & 59.0± 0.6 & 58.4± 0.5 & \textbf{64.1± 0.5} \\
Toxigen & 74.5± 0.8 & 74.0± 0.7 & \underline{74.6± 1.1} & 70.1± 0.9 & 67.6± 0.8 & \textbf{75.0± 1.1} \\
YELP & 84.0± 0.1 & 84.1± 0.2 & \textbf{84.2± 0.2} & 83.1± 0.1 & 72.8± 0.2 & \textbf{84.2± 0.1} \\
\midrule
Rank & \underline{2.7} & 2.8 & 3.0 & 5.1 & 5.5 & \textbf{1.8} \\
\bottomrule
\end{tabular}
}
\label{tab:ablation}
\end{table}

We ablate head design, initialization, and loss terms. We compare the full classification head ($|C|$ outputs) initialized from the base model against a randomly initialized classification head and a randomly initialized single-output linear head on $\vecH$ (logistic regression), all trained with BCE loss. We also retrain ALIEN after removing L2-SP, entropy regularization, or both.

\begin{itemize}
  \item \emph{Trained CLS. BCE} — Initialize the uncertainty head with $\theta_{\text{init}}$ (copied from the trained classification head), apply the entropy mapping, and train using only binary cross-entropy (BCE) on the per-sample error label.
  \item \emph{Trained CLS. BCE+L2SP} — Same initialization and entropy mapping as above; train with BCE plus L2-SP weight anchoring to $\theta_{\text{init}}$ (i.e., $\lVert \theta - \theta_{\text{init}} \rVert_2^2$).
  \item \emph{Trained CLS. BCE+Reg} — Same initialization and entropy mapping; train with BCE plus an output-consistency regularizer that keeps $U_{\text{ALIEN}}(x)$ close to the original entropy $U_{\text{Entropy}}(x)$ (i.e., $(U_{\text{ALIEN}}(x)-U_{\text{Entropy}}(x))^2$).
  \item \emph{Rand Linear. BCE} — Replace the head with a randomly initialized single-output linear layer on $z$; train with BCE to predict the per-sample error indicator.
  \item \emph{Rand CLS. BCE} — Randomly initialize a head with the same architecture as the classification head (shape of $\theta_{\text{init}}$), include the entropy mapping, and train with BCE to predict the error indicator.
\end{itemize}

Table~\ref{tab:ablation} presents the ablation results. Initializing ALIEN from the original classification head is clearly beneficial: both a random classification-style head and a single-output linear probe perform worse and have higher average ranks. This shows the value of starting from a strong pre-trained head.

The full ALIEN objective (BCE+reg+L2SP) achieves the best mean rank. Removing either regularization term hurts performance, and the entropy regularizer appears more useful than L2-SP alone, with BCE+reg outperforming BCE+L2SP ($2.5$ vs.\ $3.0$).





\section{Conclusion}                                                
\label{sec:conclusion}

We introduced ALIEN, a lightweight post-hoc uncertainty head for LoRA-tuned LLMs. Our method starts with the model’s own predictive entropy and refines it to better quantify uncertainty. By copying the trained classifier head and optimizing a three-term objective, we improve the separation between correct and incorrect predictions. ALIEN adds only ${\sim}0.002\%$ to ${\sim}0.5\%$ additional parameters, depending on whether it is applied to decoder or encoder models, and incurs negligible computational overhead during inference.

For both classification and NER problems, our method outperforms all considered modern baselines and empirically better captures aleatoric uncertainty, as the original entropy fails to do so, also achieves better calibration quality, as reflected by lower Expected Calibration Error. Ablations confirm that each loss component is necessary: using a random initialization or dropping either regularizer lowers quality, whereas the full objective yields controlled adaptation and improves quality.
Looking ahead, combining ALIEN with post-hoc calibration techniques or incorporating uncertainty-awareness during training are promising directions to further improve reliability.




\begin{credits}

\subsubsection{\discintname}
The authors have no competing interests to declare that are relevant to the content of this article.

\end{credits}

\bibliographystyle{splncs04}
\bibliography{custom}

\appendix



\end{document}